\documentclass[,12pt]{elsarticle}

\usepackage{amssymb}
\usepackage{amsmath}

\usepackage{booktabs}
\usepackage{multicol}
\usepackage{makecell}
\usepackage{subcaption}
\usepackage[font=small]{caption}
\usepackage{multirow}
\usepackage{xspace}
\usepackage{array}

\usepackage{hyperref}

\newcommand{\hsil}{CIN2$^+$\xspace}
\newcommand{\lsil}{CIN1$^-$\xspace}
\newcommand{\devtrain}{DS$\_$dev$\_$train\xspace}
\newcommand{\devval}{DS$\_$dev$\_$val\xspace}
\newcommand{\testid}{DS$\_$test$\_$id\xspace}
\newcommand{\testood}{DS$\_$test$\_$ood\xspace}

\newcommand{\DSgermany}{Test DS Germany\xspace}
\newcommand{\DSindia}{Test DS India\xspace}
\newcommand{\DScambodia}{Test DS Cambodia\xspace}
\newcommand{\DSromania}{Test DS Romania\xspace}

\newcolumntype{P}[1]{>{\centering\arraybackslash}p{#1}}

\journal{}

\begin{document}

\begin{frontmatter}



\title{Towards Global AI-Driven Cervical Cancer Screening} 


\author[1]{Thuy Nuong Tran \corref{cor1}}
\author[1]{Ömer Sümer}
\author[1]{Evangelia Christodoulou}
\author[2,3]{Lennart Nauschütte}
\author[4]{Simon Kalteis}
\author[2,3]{Martin Paulikat}
\author[5]{Esmira Pashayeva}
\author[3]{Klara Steinheuer}
\author[2]{Isabella Borges}
\author[1]{Piotr Kalinowski}
\author[2,3,4]{Hermann Bussmann}
\author[10]{Sieng Sokmney}
\author[11]{Poeung Kuong}
\author[12]{Sathiarany Vong}
\author[5]{Achim Schneider}
\author[2,3,4]{Magnus von Knebel-Doeberitz}
\author[1]{Patrick Godau \fnref{equal1}}
\author[1,6,7,8,9]{Lena Maier-Hein \fnref{equal1}}

\affiliation[1]{organization={Intelligent Medical Systems E130, German Cancer Research Center (DKFZ)},
            city={Heidelberg},
            country={Germany}}
\affiliation[2]{organization={Department
of Applied Tumor Biology, Universitätsklinikum Heidelberg},
            city={Heidelberg},
            country={Germany}}
\affiliation[3]{organization={Cooperation Unit F210, German Cancer Research Center (DKFZ)},
            city={Heidelberg},
            country={Germany}}
\affiliation[4]{organization={PAiCON GmbH},
            city={Heidelberg},
            country={Germany}}
\affiliation[5]{organization={MVZ im Fürstenberg-Karree Berlin},
            city={Berlin},
            country={Germany}}
\affiliation[6]{organization={National Center for Tumor Diseases (NCT)},
            city={Heidelberg},
            country={Germany}}
\affiliation[7]{organization={Medical Faculty, Heidelberg University},
            city={Heidelberg},
            country={Germany}}
\affiliation[8]{organization={DKFZ Heidelberg, Helmholtz Imaging},
                       country={Germany}}
\affiliation[9]{organization={Faculty of Mathematics and Computer Science, Heidelberg University},country={Germany}}
\affiliation[10]{organization={Soth Nikum District Hospital},
                           country={Cambodia}}
\affiliation[11]{organization={Siem Reap Provincial Hospital},
                           country={Cambodia}}
\affiliation[12]{organization={Ministry of Health},
                           country={Cambodia}}

\cortext[cor1]{Corresponding author: t.tran@dkfz-heidelberg.de}
\fntext[equal1]{These authors contributed equally as last authors.}

\begin{abstract}
The global elimination of cervical cancer is a key public health goal set by the World Health Organization (WHO), with screening programs reducing mortality by up to 80\%. However, access to experts and biopsy services is limited in low- to middle-income countries (LMICs). Deep learning (DL)-based algorithms offer promising support for screening, but most existing approaches have been developed and validated on private datasets from single countries.
We present the first DL-based approach to cervical cancer screening validated on data from multiple countries. Technically, we phrase the problem of detecting and classifying lesions in colposcopy images as a multi-task learning problem, in which we simultaneously perform image-level classification and lesion segmentation. Our model was trained on a private data set of acid stain colposcopy images with manually generated lesion segmentation masks and corresponding histopathological results, employing extensive data augmentation to address image variability.
In an in-distribution validation with pathology results serving as ground truth, our algorithm outperformed medical experts (Balanced Accuracy: 0.68 vs 0.64) in CIN1- (Cervical intraepithelial neoplasia grade 1 or lower) versus CIN2+ (grade 2 or higher) classification. External validation on four colposcopy data sets from four countries featuring radical differences in prevalence and patient characteristics yielded superior performance of our method compared to baseline methods. Performance variability across countries was high with AUC values ranging from 0.54 - 0.80. Overall, algorithm performance varied with age, transformation zone (cervical area most prone to lesion development), presence of comorbidities and pathognomonic signs, with comorbidities having by far the largest negative effect. Future work should focus on improving model robustness and generalizability. 

\end{abstract}







\end{frontmatter}



\section{Introduction}
\label{sec:introduction}
The World Health Organization (WHO) has identified the global elimination of cervical cancer as a major public health priority. Regular screening is estimated to reduce cervical cancer mortality by up to 80\%~\cite{whomort}. A central challenge in screening programs is the accurate classification of cervical intraepithelial neoplasia (CIN), particularly the differentiation between \lsil (normal or grade 1) and \hsil (grade 2 or higher), which informs treatment decisions. This classification task is inherently difficult and highly dependent on the skill and experience of the colposcopist.

Reported diagnostic accuracy for differentiating \lsil from \hsil varies substantially, with values ranging from approximately 50\%, which is only marginally above chance level, to about 70\%~\cite{chen2023application, bai2022assessing, wei2022improving}.
These limitations are exacerbated in low- to middle-income countries (LMICs), where access to expert colposcopists and biopsy services is often restricted. Computer-aided analysis offers a promising avenue to support both the training of gynecologists and the screening process itself, particularly in resource-constrained environments.\\

\begin{figure}[t]
    \begin{minipage}{0.74\textwidth}
        \centering
        \includegraphics[height=0.18\textheight]{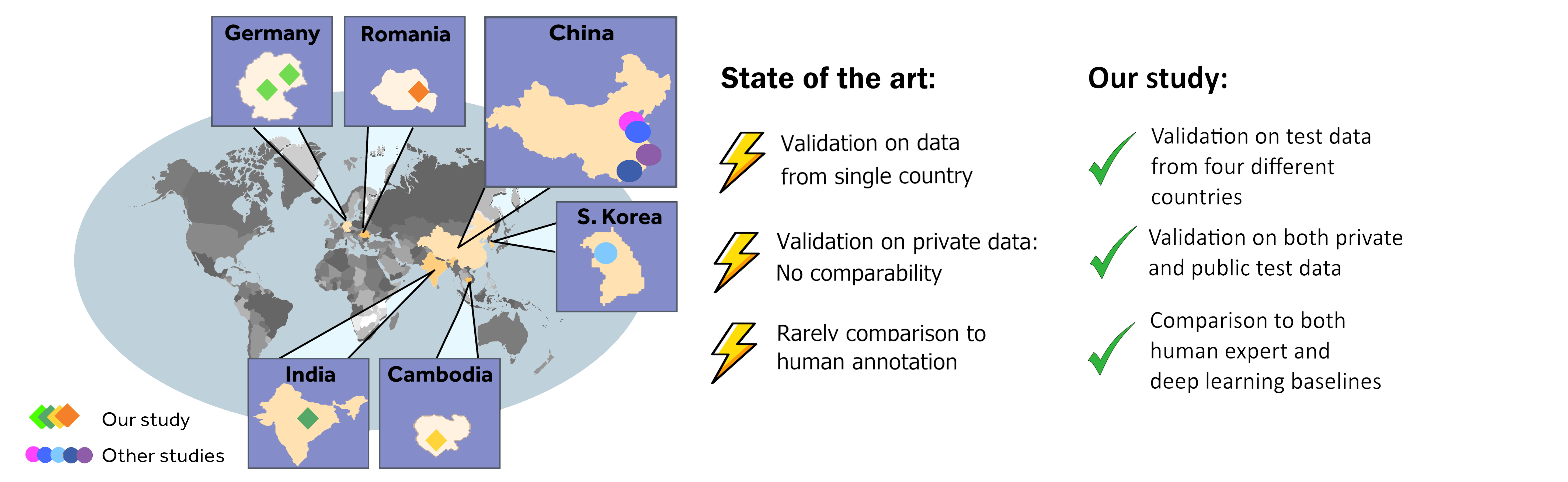}\\
        {\footnotesize \textbf{(a)}}
    \end{minipage}
    \begin{minipage}{0.24\textwidth}
        \centering
        \includegraphics[height=0.18\textheight]{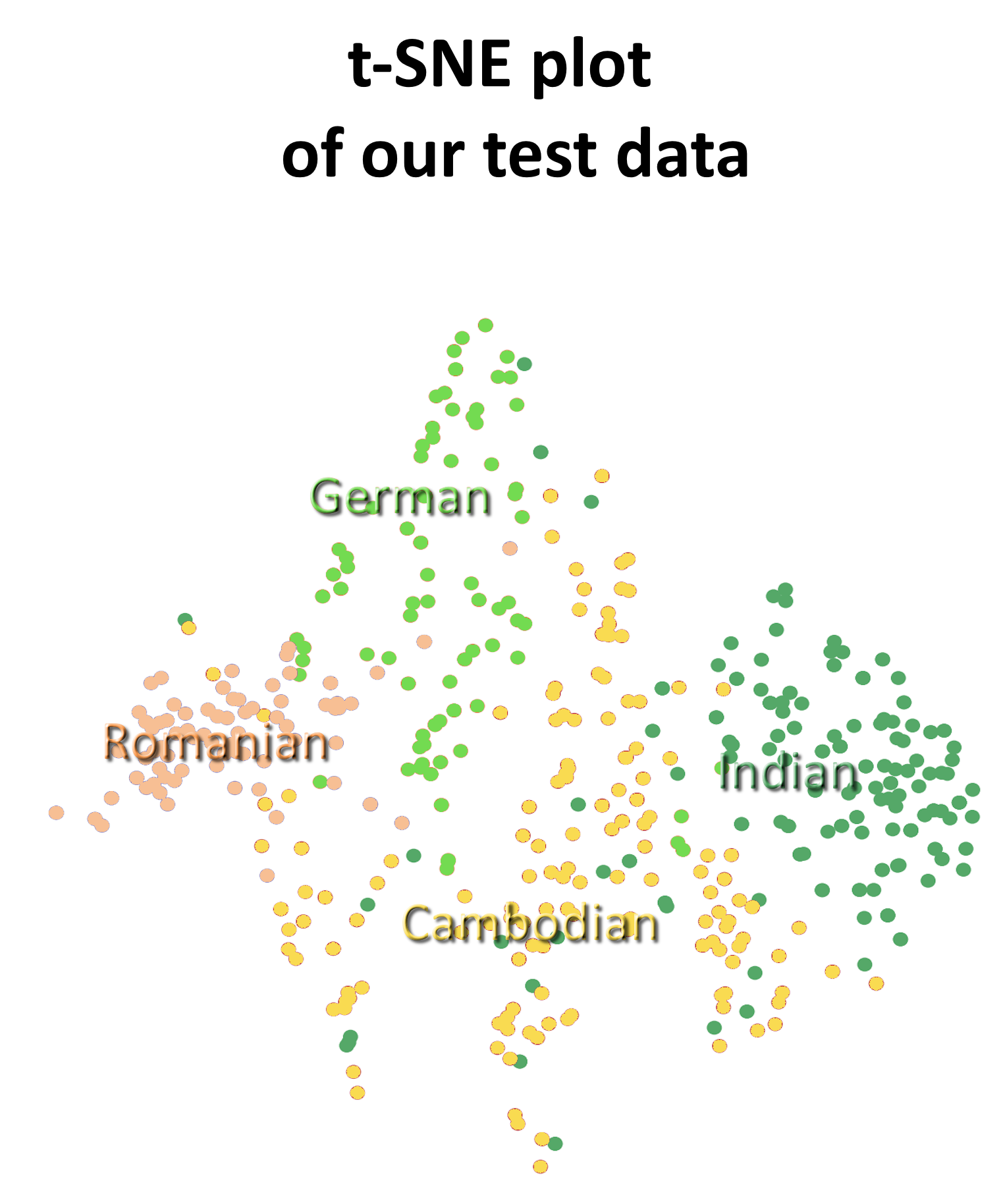}
        {\footnotesize \textbf{(b)}}
    \end{minipage}
    \caption{\textbf{Previous approaches to DL-based cervical cancer screening have been developed and validated on private datasets from single countries despite high image variability.} (a) Distribution of data from our study (diamonds) and previous studies (circles). (b) 2D visualization of our test data, color-coded by site. }
    \label{fig:worldmap}
    
\end{figure}

\begin{figure}[h]
    \begin{minipage}{0.70\textwidth}
        \centering
        \includegraphics[height=0.22\textheight]{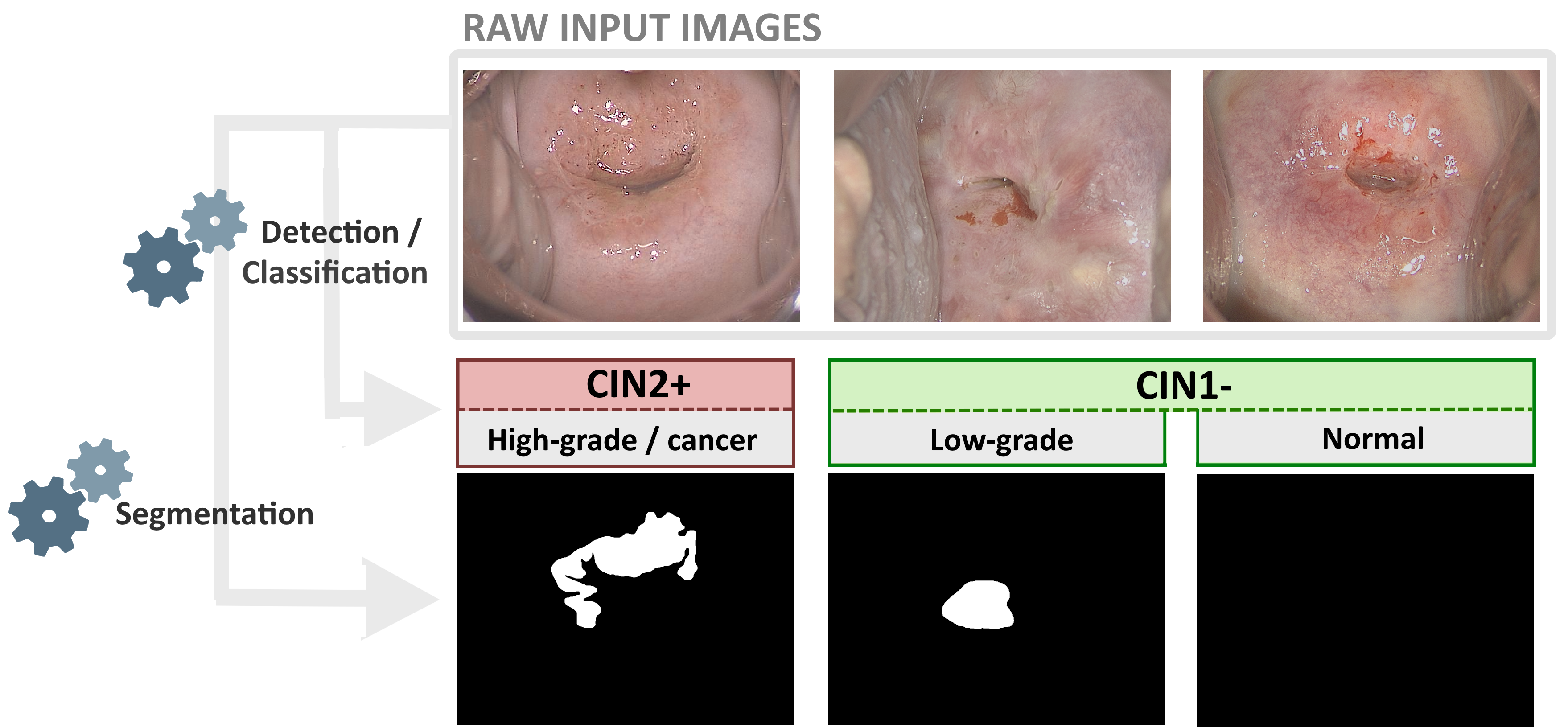}\\
        {\footnotesize \textbf{(a)}}
    \end{minipage}
    \begin{minipage}{0.20\textwidth}
        \centering
        \includegraphics[height=0.22\textheight]{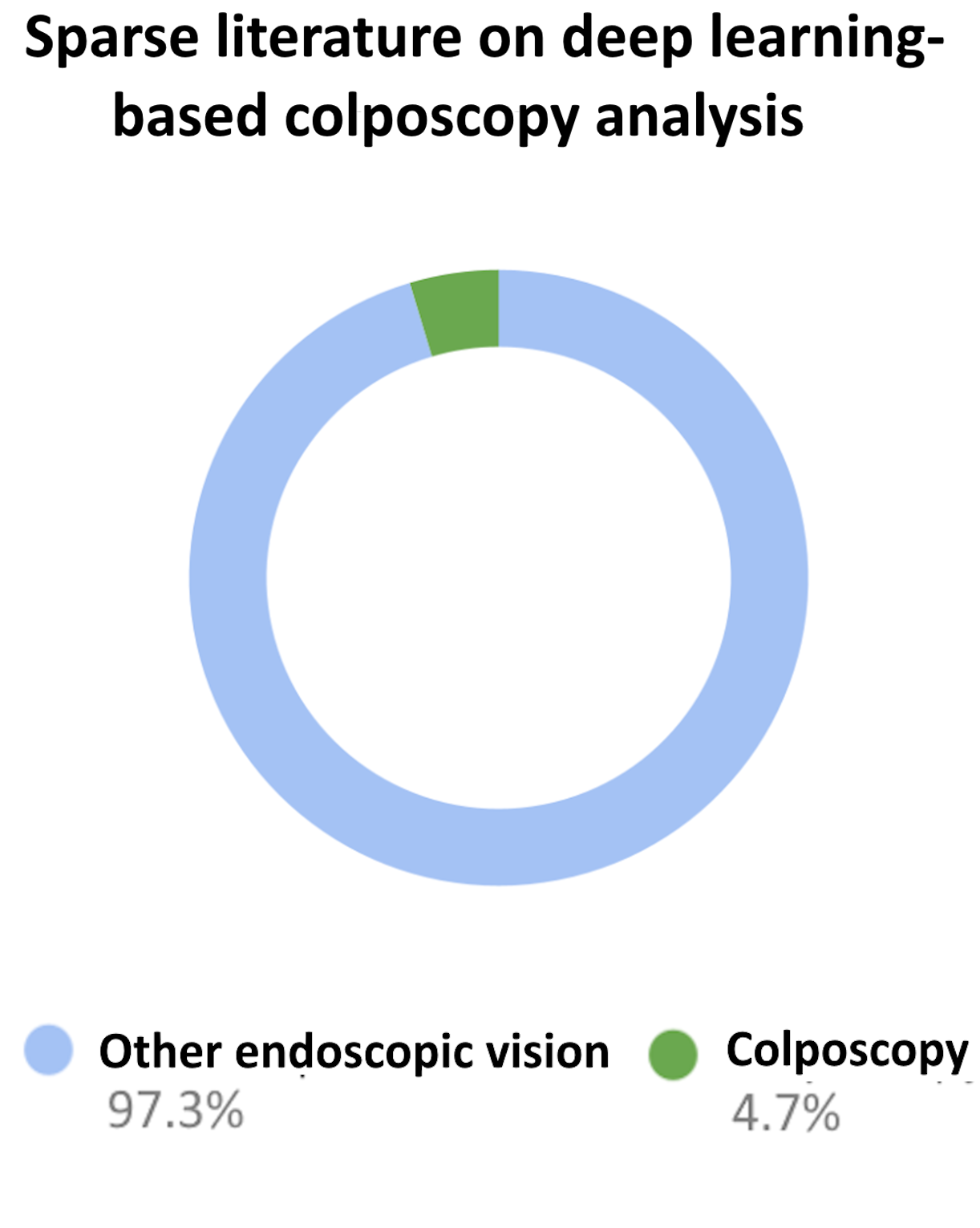}\\
        {\footnotesize \textbf{(b)}}
    \end{minipage}
     \caption{\textbf{Lesion classification and segmentation in colposcopy is a highly difficult and under-explored task.} (a) Automated detection of cervical lesions can serve as assistance for clinicians during cancer screenings. However, the classification of cervical lesions into \lsil cases, which include normal and low-grade lesions, versus \hsil lesions, which include high-grade lesions and cancer, a task that is relevant to determine the need for therapy, is considered difficult even among experts~\cite{chen2023application, bai2022assessing, wei2022improving}. (b) Despite the high clinical relevance, DL-based publications for colposcopy are underrepresented in the computer vision community. A structured literature search on PubMed and Google Scholar was conducted to identify relevant studies at the intersection of colposcopy, deep learning, and validation. The analysis revealed that previous work on AI-based lesion classification and performance assessment was limited in scope and generalizability. }
    \label{fig:difficult}
    
\end{figure}

Despite this potential, prior studies have been limited in scope, relying on data from single countries or institutions, as illustrated in Fig.~\ref{fig:worldmap}. In addition, existing deep learning (DL)-based approaches in this domain remain sparse, reflecting the difficulty of the task, as shown in Fig.~\ref{fig:difficult}(b). Li et al.~\cite{li2023segmentation} employed a Deeplabv3+ model, evaluating it on data from a single hospital in China. Yu et al.~\cite{yu2022segmentation} proposed CLSNet, which combines a region proposal network with EfficientNet-B3, also validated on data from one hospital in China. Cho et al.~\cite{cho2020classification} investigated Inception-ResNet-v2 and ResNet-152, using data from three hospitals in Seoul, though without incorporating multi-country validation or out-of-distribution testing. Other studies, such as those by Yan et al.~\cite{yan2021multi} and Sha et al.~\cite{sha2024cervifusionnet}, relied on multimodal inputs, including saline or iodine images and patient metadata. These modalities are often impractical for real-time deployment due to acquisition constraints.\\

Technically, all prior work formulated the problem as a single-task supervised learning task, without exploring multi-task learning frameworks combined with the use of augmentation strategies during training or inference to address intra- and inter-patient variability. Although high variability across data sources is evident, as illustrated in Fig. \ref{fig:worldmap}(b), no existing study has conducted validation across multiple countries. An additional limitation is the lack of stratified failure analysis using available metadata, which remains an underutilized resource. While metadata such as patient age or transformation zone type has occasionally been used as auxiliary input to DL models\cite{yuan2020application}, it has rarely been leveraged to investigate performance breakdowns across subpopulations. For example, Li et al.~\cite{li2024evaluation} stratified diagnostic accuracy of human colposcopists by age group but did not incorporate metadata-driven failure analysis within a DL framework, nor did their study consider broader patient characteristics. \\

We address these gaps in the literature with the following contributions: 
\textbf{New method:} We present a novel DL-based approach for cervical cancer screening that formulates the problem as a multi-task learning task, using segmentation masks and extensive data augmentation during both training and testing to address image variability. \\
\textbf{Large-scale validation: }We perform large-scale validation using public and private datasets from four countries(see  Fig. \ref{fig:worldmap}), demonstrating superior performance compared to human annotators and established baseline methods. \\
\textbf{Identification of failure cases:} We conduct detailed failure analysis by stratifying model performance with respect to patient-specific variables, including age, comorbid pathologies, transformation zone type, and pathognomonic signs, as illustrated in Fig. \ref{fig:pathosigns}.\\

\newpage
\section{Materials and Methods}
\label{sec:material}
Section \ref{sec:data} introduces the datasets utilized for model development and evaluation. Section \ref{sec:multi-task} presents the multi-task learning approach, including core concepts, implementation, and ensembling. Section \ref{sec:experiment-design} describes the experimental design used to evaluate performance, compare to previous methods, and analyze failures cases using metadata stratification.

\subsection{Data}
\label{sec:data}
In the following, we provide details on data characteristics, our data splitting strategy and curation process.

\subsubsection{Dataset Overview}
\label{sec:data-overview}
We utilized colposcopic image data from four countries to ensure diverse representation in ethnicity, imaging conditions and device characteristics. Privately acquired datasets from Cambodia and Germany were complemented by two public datasets: the Indian Atlas of Colposcopy\cite{iarc-dataset} from the WHO and the Romanian AnnoCerv dataset by Socolov et al.~\cite{socolov2023annocerv}. 
Ethical approval for this study was granted by the Ethics Committee of the Medical Faculty of Heidelberg (reference number pending). All data were anonymized, and informed consent was obtained where applicable, following the Declaration of Helsinki. The inclusion of data across multiple countries introduces substantial variation in imaging quality, lighting, and sensor types, making cross-site generalization a critical research question. Detailed dataset statistics are provided in Table \ref{tab:data_statistics}.

\begin{table}[!tbp]
    \centering
    \begin{tabular}{P{0.6cm}P{1.3cm}P{1.3cm}P{1cm}P{1cm}P{1cm}P{1cm}}
        \toprule
       & \multicolumn{2}{c}{\shortstack{Training/\\Validation}}
& \multicolumn{1}{c}{\shortstack{Test DS \\ Germany}} 
& \multicolumn{1}{c}{\shortstack{Test DS \\ India}} 
& \multicolumn{1}{c}{\shortstack{Test DS \\ Cambodia}} 
& \multicolumn{1}{c}{\shortstack{Test DS \\ Romania}} \\
\cmidrule(lr){2-3} \cmidrule(lr){4-4} \cmidrule(lr){5-7} 
        & \shortstack{\small DS\_dev\_\\ \small train} & \shortstack{\small DS\_dev\_\\ \small val} & \small DS\_test\_id &  \multicolumn{3}{c}{\small DS\_test\_ood}   \\
        \midrule
        {\textbf{$N$}} & 696 & 174 & 177 & 197 & 165 & 100 \\[5pt]
      {$\pi$} & \multicolumn{2}{c}{0.64} & 0.63 & 0.23 & 0.2 & 0.37  \\
        \bottomrule
    \end{tabular}
    \caption{\textbf{Training and test data statistics.} Training and validation consist of data sourced from a private German database, as well as the publicly available IARC Colposcopy Imagebank~\cite{iarc-colpobank} from India. An in-distribution (ID) German dataset is reserved for testing (\DSgermany). Furthermore, we acquired images from Cambodia as an out-of-distribution(OOD) test set (\DScambodia). Lastly, we use two additional OOD datasets for testing, namely publicly available IARC VIA Imagebank ~\cite{iarc-via}(\DSindia) from a separate Indian hospital and AnnoCerv from Romania~\cite{socolov2023annocerv} (\DSromania). Here, $N$ denotes the number of samples and $\pi$ the prevalence of the positive class.
}
    \label{tab:data_statistics}
\end{table}

\subsubsection{Data Splitting}
To prevent data leakage and overfitting, we employed a strict data splitting protocol that separates development and evaluation phases. The development dataset included training and validation subsets, while the test dataset remained untouched during model development.
The development dataset was split into 80\% training and 20\% validation. The training subset (\devtrain) consisted of 696 images: 539 from the German dataset and 157 from the IARC Colposcopy Imagebank\cite{iarc-colpobank}. The validation subset (\devval) included 174 images: 135 from Germany and 39 from IARC, and was used for hyperparameter tuning and ablation studies.
The test dataset comprised two components: in-distribution and out-of-distribution samples. The in-distribution test set (\testid) contained 177 German images. The out-of-distribution test set (\testood) included 197 Indian cases from the IARC VIA ImageBank\cite{iarc-via}, 165 Cambodian cases, and 100 Romanian images from AnnoCerv\cite{socolov2023annocerv}, totaling 462 external samples.

\subsubsection{Dataset Curation}
\label{sec:data-curation}
\begin{figure}[h]
    \centering
    \includegraphics[width=\textwidth]{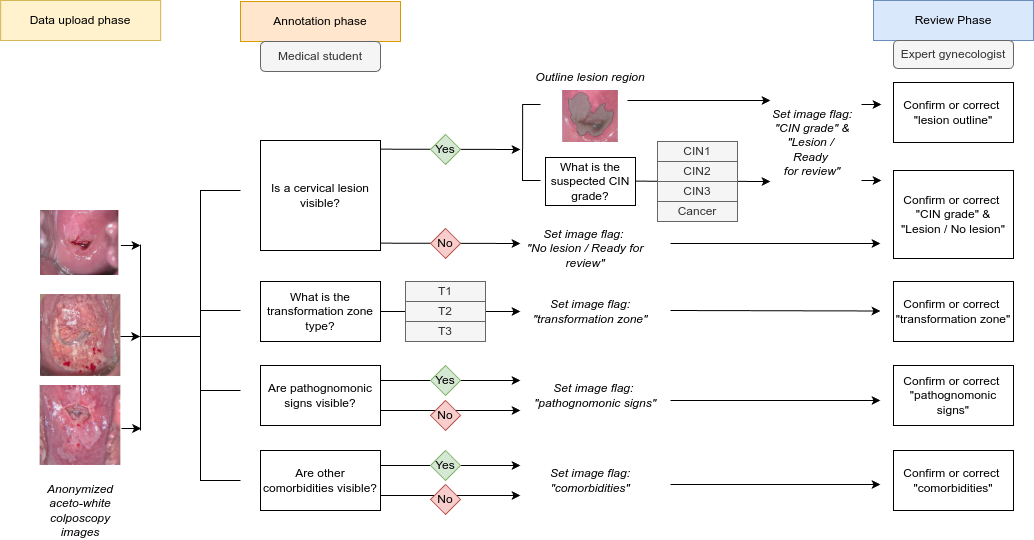}
    \caption{\textbf{Annotation protocol presented to clinicians for privately acquired data.} Anonymized aceto-white colposcopy images were presented to medical students for localization of lesions and they were asked to grade lesion severity to the best of their knowledge. They were further asked to label transformation zone, comorbidity and pathognomonic sign presence. After these steps, the images were tagged as ready for review for the expert gynecologist. The expert confirmed or corrected the findings by approving or adjusting the annotations.
}
    \label{fig:annotatio-protocol}
\end{figure}

Colposcopic screening is used to detect and grade cervical intraepithelial neoplasia (CIN), which can be categorized as normal, CIN1, CIN2, CIN3, or invasive cancer\cite{mello2023cervical}. 
For private datasets, one aceto-white image per patient was selected based on optimal cervix visibility. Histopathological biopsy results served as reference during training and evaluation. For multi-task training, expert-provided segmentation masks and binary class annotations were obtained to serve as human baselines. Annotation followed a two-stage protocol: initial lesion delineation and grading by two trained medical students, followed by expert review for label confirmation or correction. See Fig. \ref{fig:annotatio-protocol} for full details.
For IARC and AnnoCerv datasets, one acid-stain image per case was selected, with labels extracted from corresponding metadata.
To enable stratified failure analysis, we curated metadata for \devtrain (Germany), including age, transformation zone type, comorbidities, and presence of pathognomonic signs. Age was binned by decade, starting from $<$30 up to $>$60. Transformation zones were categorized as: Type 1 (fully visible), Type 2 (partially receded), and Type 3 (mostly hidden), as shown in Fig. \ref{fig:pathosigns}(a).
Documented comorbidities included condylomata acuminata, ectopy, atrophy, and polyps (see Fig. \ref{fig:pathosigns}(b)). These conditions can obscure lesion boundaries or mimic pathological appearances. Condylomata, caused by low-risk HPV, and spontaneous changes like ectopy or atrophy introduce diagnostic complexity.
Pathognomonic signs, indicative of high-grade lesions, were also recorded. Our study data contained cases exhibiting the following pathognomonic signs: Rag signs (epithelial peeling), inner border signs (distinct, well-defined acetowhite line marking the inner margin of a cervical lesion), ridge signs (raised or thickened epithelial ridges), cuffed gland openings (glands surrounded by acetowhite epithelium), fine or coarse punctation (dot-like vascular patterns within the lesion) and fine or coarse mosaic (vascular pattern with tile-like appearance). Visualizations are shown in Fig. \ref{fig:pathosigns}(c).

\subsection{Multi-task Learning Approach}
\label{sec:multi-task}
The clinical objective is to assess whether therapy is required by detecting cervical lesions and distinguishing high-grade from low-grade findings. Since only CIN2+ cases require treatment, we framed the task as binary classification problem (see Fig. \ref{fig:difficult}(a)): \lsil (normal and CIN1) versus \hsil (CIN2, CIN3, and cancer). 

\subsubsection{Concept}
To achieve this, our approach builds upon three core design components. First, we formulate cervical cancer screening as a multi-task learning problem. As shown in Fig. \ref{fig:difficult}(a), lesion detection and classification are challenging tasks. To improve feature representation, we incorporate lesion segmentation masks and jointly train for lesion classification and segmentation. Second, given the high visual variability in colposcopy images, illustrated in Fig. \ref{fig:worldmap}(b), we apply extensive data augmentation during training and testing to enhance model robustness. Third, we adopt a model ensembling strategy, following the methodology outlined by Eisenmann et al.~\cite{eisenmann2023winner}, to improve prediction stability and generalization.

\subsubsection{Implementation Details}
The proposed architecture is based on EfficientNet-B4~\cite{tan2019efficientnet} with dual heads: a classification head that distinguishes \lsil from \hsil cases, and a segmentation head that learns the region of interest. During training, we applied standard augmentations including rotation, flipping, and translation, as well as complex transformations such as random-resized-crop, color transfer, and color jitter, followed by resizing to 320x320 px and normalization. For heavy augmentation, we employed FancyPCA to perturb color channels, Gaussian noise to simulate device specific acquisition variability, and randomized Cutout to obscure up to 15\% of the image, each applied with 50\% probability. 
Test-time augmentation involved randomly applying horizontal or vertical flips and center-cropping, and finally resizing and normalization. Each image underwent four augmented evaluations, and predictions were averaged. 
Model ensembling was performed using five-fold cross-validation on the development dataset (\devtrain + \devval). Each fold was trained using the design decision that achieved the best validation performance. Final predictions were computed by averaging class probabilities across all five models.

\begin{figure}[!htbp]
    \centering
    \begin{subfigure}{\linewidth}
        \centering
        \includegraphics[width=0.9\linewidth]{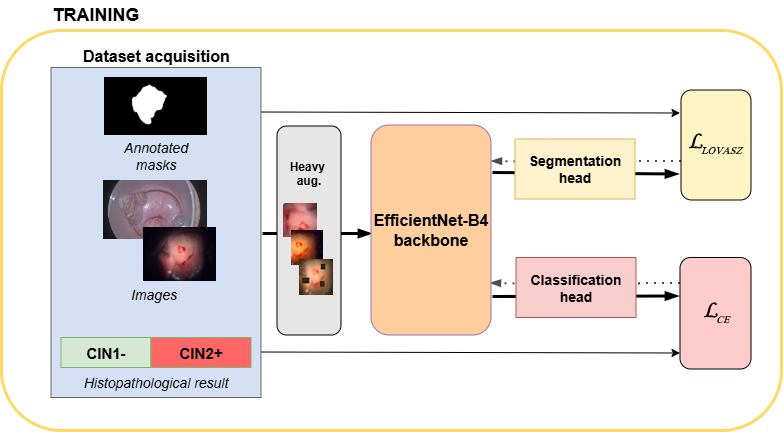}
    \end{subfigure}

    \vspace{1em} 

    \begin{subfigure}{\linewidth}
        \centering
        \includegraphics[width=0.9\linewidth]{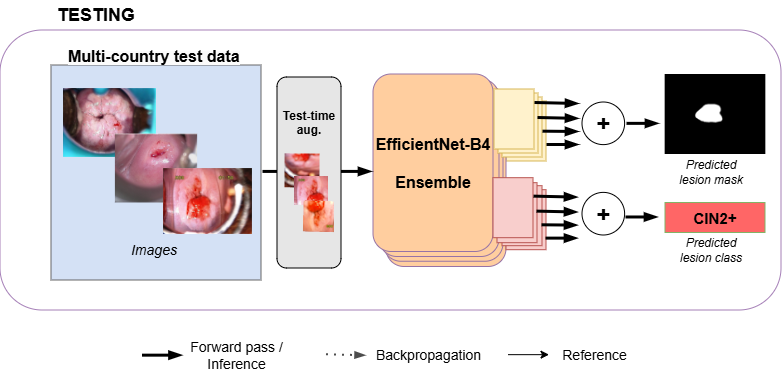}
    \end{subfigure}

    \caption{ \textbf{Pipeline overview. }(Training) The problem of cervical cancer screening was phrased as a multi-task learning problem using histopathology results and human lesion segmentation as reference data for training. An EfficientNet-B4 served as backbone for a classification head for \lsil/\hsil lesion classification. A segmentation head was employed to capture spatial information that improves feature representations in the latent space. To increase robustness, we made heavy use of augmentations during training (Heavy aug.). (Testing) To account for high image variability, test-time augmentation (Test-time aug.) was applied to an unseen image using rotation, scaling and center-crop. An ensemble of five trained models was used for inference. The outputs of the ensemble are averaged at the plus node for the final prediction.}
    \label{fig:pipeline}
\end{figure}

\subsection{Experimental Design}
\label{sec:experiment-design}
Our experiments address four key questions, Q1–Q4, designed to evaluate model performance, compare to existing methods, assess design decisions, and understand failure modes.
\newpage
\textbf{Q1: How does our method perform compared to human annotators?}
 We evaluated model performance against human baseline annotations described in section \ref{sec:data-curation}, using histopathological diagnosis as reference. Following Maier-Hein et al.~\cite{maier2024metrics}, we report balanced accuracy (BA) to account for class imbalance. Sensitivity is emphasized due to its clinical relevance in identifying high-grade lesions. Specificity is also reported to assess the model’s ability to avoid false positives, which is critical in minimizing unnecessary interventions. To further capture diagnostic reliability, we report the F1-score for the positive class, which balances sensitivity and precision. Confidence intervals were estimated using 1,000 bootstrap resamples, as performance variability is critical for assessing clinical translation in biomedical imaging AI, following Christodoulou et al.~\cite{christodoulou2024confidence}.\\
 
\textbf{Q2: How does our proposed method perform compared to previous methods?}
 We closely re-implemented two architectures from prior work for comparison: ResNet-152 (Cho et al.~\cite{cho2020classification}) and DeepLabv3+ (Li et al.~\cite{li2023segmentation}), adapting their classification heads for binary classification.
 In accordance with the Metrics Reloaded framework, we report a target value-based metric Specificity@Sensitivity=0.9 (Spec@Sens=0.9), prioritizing sensitivity for high-grade lesion detection, even at the cost of increased false positives. Area under the ROC curve (AUC) is provided as a threshold-independent metric, commonly used in classification tasks to reflect the model's ability to discriminate between classes across all decision thresholds. Finally, BA is included for comparability across datasets with different prevalence rates.\\
 
\textbf{Q3: What are the most important design decisions?}
 To assess the impact of each component of our pipeline, we conducted ablation studies on the validation set (\devval) and evaluated their generalization on our four country-specific test sets (\testid + \testood). Starting from a baseline EfficientNet-B4 pretrained on ImageNet~\cite{5206848}, we sequentially introduced multi-task learning, heavy training augmentations, test-time augmentation, and ensembling. Performance gains were evaluated at each stage.
\\

\begin{figure}[!htbp]
    \centering
    \begin{subfigure}[b]{\textwidth}
        \centering
        \includegraphics[width=0.65\linewidth]{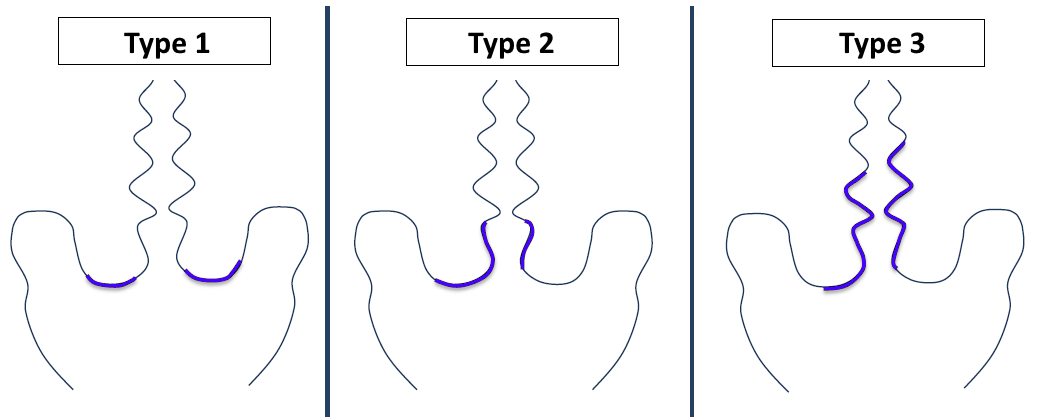}
    \end{subfigure}
    {\footnotesize \textbf{(a)}}
    \par\vspace{0.3cm}
    \begin{minipage}[b]{0.23\linewidth}
        \centering
        \includegraphics[width=\linewidth]{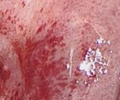}
        \caption*{Atrophy} 
    \end{minipage}
    \hfill
    \begin{minipage}[b]{0.23\linewidth}
        \centering
        \includegraphics[width=\linewidth]{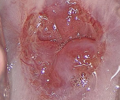}
        \caption*{Ectopy}
    \end{minipage}
    \hfill
    \begin{minipage}[b]{0.23\linewidth}
        \centering
        \includegraphics[width=\linewidth]{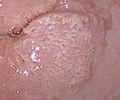}
        \caption*{Condyloma }
    \end{minipage}
    \hfill
    \begin{minipage}[b]{0.23\linewidth}
        \centering
        \includegraphics[width=\linewidth]{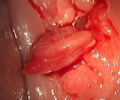}
        \caption*{Polyp}
    \end{minipage}
    \par
    {\footnotesize \textbf{(b)}}
    \par \vspace{0.3cm}
    \begin{subfigure}[b]{\textwidth}
    \centering
    \begin{minipage}[b]{0.23\linewidth}
        \centering
        \includegraphics[width=\linewidth]{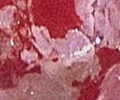}
        \caption*{Rag sign}
    \end{minipage}
    \hfill
    \begin{minipage}[b]{0.23\linewidth}
        \centering
        \includegraphics[width=\linewidth]{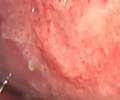}
        \caption*{Inner border sign}
    \end{minipage}
    \hfill
    \begin{minipage}[b]{0.23\linewidth}
        \centering
        \includegraphics[width=\linewidth]{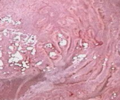}
        \caption*{Ridge sign}
    \end{minipage}
    \hfill
    \begin{minipage}[b]{0.23\linewidth}
        \centering
        \includegraphics[width=\linewidth]{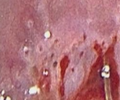}
        \caption*{Cuffed crypt}
    \end{minipage}

    \par\vspace{6pt} 

    \begin{minipage}[b]{0.23\linewidth}
        \centering
        \includegraphics[width=\linewidth]{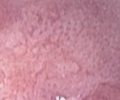}
        \caption*{Fine punctation}
    \end{minipage}
    \hfill
    \begin{minipage}[b]{0.23\linewidth}
        \centering
        \includegraphics[width=\linewidth]{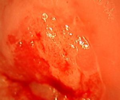}
        \caption*{Coarse punctation}
    \end{minipage}
    \hfill
    \begin{minipage}[b]{0.23\linewidth}
        \centering
        \includegraphics[width=\linewidth]{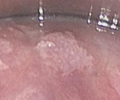}
        \caption*{Fine mosaic}
    \end{minipage}
    \hfill
    \begin{minipage}[b]{0.23\linewidth}
        \centering
        \includegraphics[width=\linewidth]{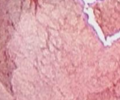}
        \caption*{Coarse mosaic}
    \end{minipage}
    \par
    {\footnotesize \textbf{(c)}}
\end{subfigure}
    \caption{\textbf{Patient marker visualization for failure analysis stratification. }(a) Transformation zone: Region at the squamocolumnar junction where cervical lesions typically arise. Transformation zone is fully visible for T1 and increasingly recedes inwards the cervix canal, making it partially visible for T2 and mostly or entirely hidden for T3.  (b) Other pathology: Illustrations of comorbidities featured in our study. (c) Pathognomonic signs: Visualization of disease-specific diagnostic indicators that were present in our study.}
    \label{fig:pathosigns}
\end{figure}

\textbf{Q4: What characterizes images on which our method fails?}
To identify systematic failure patterns, we stratified performance by the
four metadata categories outlined in section \ref{sec:data-curation}: patient age, transformation zone type, comorbidities and presence of pathognomonic signs. A representative visualization of selected patient markers is shown in Fig. \ref{fig:pathosigns}.

\newpage

\section{Results}
\label{sec:results}

\subsection{Q1: How does our method perform compared to human annotators?}
\begin{figure}[!tbp]
    \centering

    \begin{minipage}{0.63\textwidth}
        \centering
        \includegraphics[trim=1cm 3.7cm 2cm 1.5cm, clip, width=\linewidth]{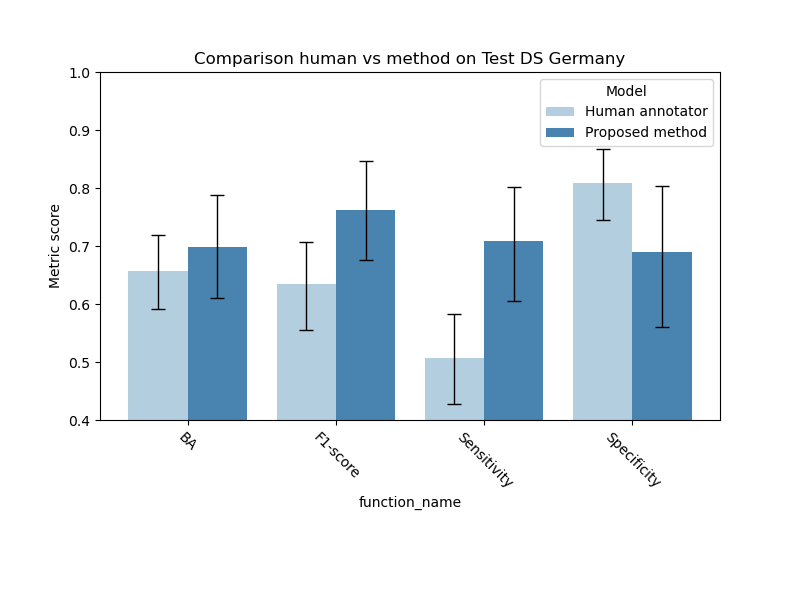}
        {\footnotesize \textbf{(a)}}
    \end{minipage}
    \begin{minipage}{0.36\textwidth}
        \centering
        \includegraphics[width=\linewidth]{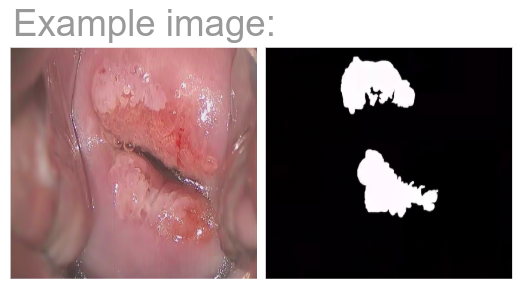}\\
        \vspace{0.3cm}
        \includegraphics[width=\linewidth]{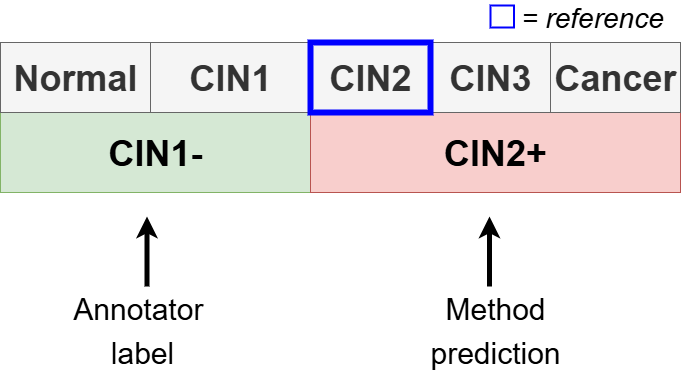}\\
        {\footnotesize \textbf{(b)}}
    \end{minipage}
    \caption{\textbf{Our model outperforms human experts using histopathology data as ground truth.}(a)  Balanced Accuracy (BA), F1-score, sensitivity, and specificity are shown for human annotators compared to our proposed method for in-distribution test data (\DSgermany). Error bars represent the 95\% bootstrap-based confidence interval. Our method shows superiority compared to the human annotator in the BA, F1-score and sensitivity metric. (b) Example image and respective segmentation mask of a lesion misclassified by a human annotator. }
    \label{fig:human_vs_model}
\end{figure}

 As shown in Fig. \ref{fig:human_vs_model}, our method outperforms human annotators in terms of overall diagnostic performance. It achieves a 6.3 percentage points (\,pp) increase in balanced accuracy, as well as higher F1-score and sensitivity. Notably, expert annotations show a low sensitivity score of 0.51 compared to our method which achieves a sensitivity of 0.71. Human annotators outperform our method in specificity, but the specificity values remain high at 0.81 for human and 0.69 for method specificity, respectively.

\subsection{Q2: How does our method compare to previous methods?}
\begin{figure}[!tbp]
    \centering
    \begin{subfigure}{0.48\textwidth}
        \centering
        \includegraphics[trim=15pt 15pt 25pt 15pt, clip, width=\linewidth]{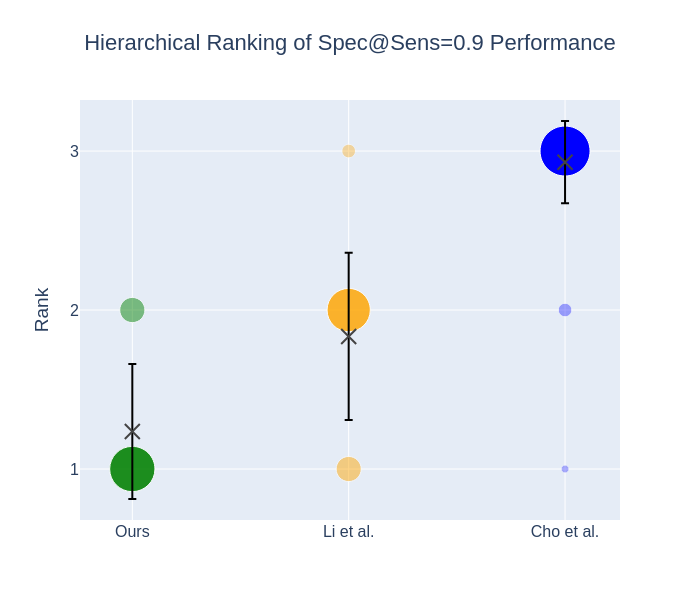}
        \label{fig:rank-a}
    \end{subfigure}
    \begin{subfigure}{0.48\textwidth}
        \centering
        \includegraphics[trim=15pt 15pt 25pt 15pt, clip, width=\linewidth]{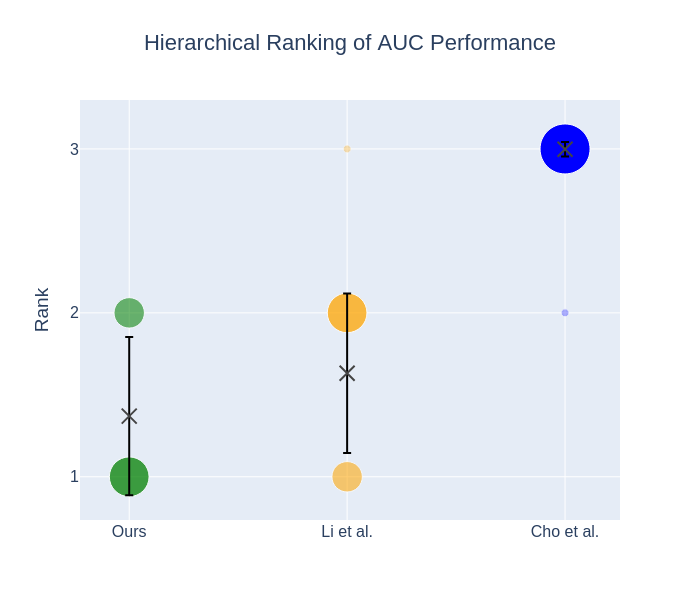}
        \label{fig:rank-b}
    \end{subfigure}
    \caption{\textbf{Our method outperforms the baseline methods across different metrics.} A Blob plot representing model ranking was created using hierarchical bootstrapping. First, the four country test sets were sampled with replacement. Next, AUC and Spec@Sens=0.9 were calculated for a bootstrapped version of that test set. The models are then ranked. Finally, this procedure is repeated 1000 times to receive a distribution of ranking scores. Blob size shows frequency of rank across the 1000 bootstrap samples. X marks the mean rank and whiskers represent the standard deviation. Plot is inspired by~\cite{wiesenfarth2021methods}.}
    \label{fig:ranking_blob}
\end{figure}

Across all country-specific test sets, all methods experienced performance degradation on out-of-distribution data. Our approach consistently outperforms both baselines  in terms of Spec@Sens=0.9. 
For {\DSgermany} and {\DSindia}, our method achieves the highest AUC, outperforming the next-best model by 21.3\,pp and 12.7\,pp, respectively. While DeepLabv3+ achieves higher AUC and balanced accuracy on the {\DScambodia} and {\DSromania}, it shows lower specificity than our method. Quantitative results are summarized in Table \ref{tab:performance}. Aggregated rankings across all countries confirm that our method demonstrates the best overall performance among all evaluated models, as shown in Fig. \ref{fig:ranking_blob}.

\begin{table}[!htbp]
    \centering
    
    
        
    
    {\small
    \begin{tabular}{lcccc}
        \toprule
        \textbf{Test Set} & \textbf{Model} & \shortstack{\textbf{Spec@Sens=0.9}  \\ (95\% CI)}& \shortstack{\textbf{AUC} \\ (95\% CI)} & \shortstack{\textbf{BA} \\ (95\% CI)} \\
        \midrule
        \multirow{3}{*}{\shortstack{Test DS \\ Germany}} 
        & Ours & 0.28 (0.13--0.43) & 0.74 (0.66--0.81) & 0.70 (0.62--0.77) \\
        & Cho et al.~\cite{cho2020classification} & 0.17 (0.06--0.31) & 0.59 (0.49--0.67) & 0.55 (0.50--0.60) \\
        & Li et al.~\cite{li2023segmentation}) & 0.08 (0.00--0.19) & 0.61 (0.53--0.69) & 0.51 (0.45--0.58) \\
        \midrule
        \multirow{3}{*}{\shortstack{Test DS \\ India}}
        & Ours & 0.38 (0.12--0.61) & 0.80 (0.71--0.88) & 0.70 (0.62--0.78) \\
        & Cho et al.~\cite{cho2020classification} & 0.11 (0.04--0.23) & 0.52 (0.42--0.62) & 0.50 (0.49--0.50) \\
        & Li et al.~\cite{li2023segmentation}) & 0.36 (0.22--0.52) & 0.71 (0.62--0.80) & 0.64 (0.56--0.72) \\
        \midrule
        \multirow{3}{*}{\shortstack{Test DS \\ Cambodia}}
        & Ours & 0.28 (0.15--0.43) & 0.60 (0.51--0.74) & 0.52 (0.45--0.59) \\
        & Cho et al.~\cite{cho2020classification} & 0.02 (0.00--0.07) & 0.44 (0.32--0.56) & 0.52 (0.49--0.57) \\
        & Li et al.~\cite{li2023segmentation}) & 0.27 (0.12--0.54) & 0.68 (0.58--0.79) & 0.62 (0.53--0.72) \\
        \midrule
        \multirow{3}{*}{\shortstack{Test DS \\ Romania}}
        & Ours & 0.20 (0.06--0.32) & 0.54 (0.41--0.65) & 0.51 (0.40--0.61) \\
        & Cho et al.~\cite{cho2020classification} & 0.08 (0.00--0.33) & 0.47 (0.35--0.60) & 0.45 (0.37--0.55) \\
        & Li et al.~\cite{li2023segmentation}) & 0.18 (0.04--0.41) & 0.59 (0.47--0.71) & 0.54 (0.46--0.62) \\
        \bottomrule
    \end{tabular}
    }

    \caption{\textbf{Test set performance for various metrics across all test sets. }We report Specificity@Sensitivity=0.90, AUC, and Balanced accuracy (BA) for our method and two baseline models, across all four test sets. Our method showcases the highest Spec@Sens=0.9 across all test sets. For \DSgermany and \DSindia, our method achieves superior performance compared to the baseline. For \DScambodia and \DSromania, the baseline method DeepLabV3 scores higher in AUC, but shows a low specificity at high sensitivity level (Specificity@Sensitivity=0.9).}

    \label{tab:performance}
\end{table}

\subsection{Q3: What are the most important design decisions?}
\label{sec:ablation}
\begin{figure}[!tbp]
    \centering

    \begin{subfigure}{0.48\textwidth}
        \centering
        \includegraphics[trim=1.5cm 3.1cm 2.7cm 1.6cm, clip, width=\linewidth]{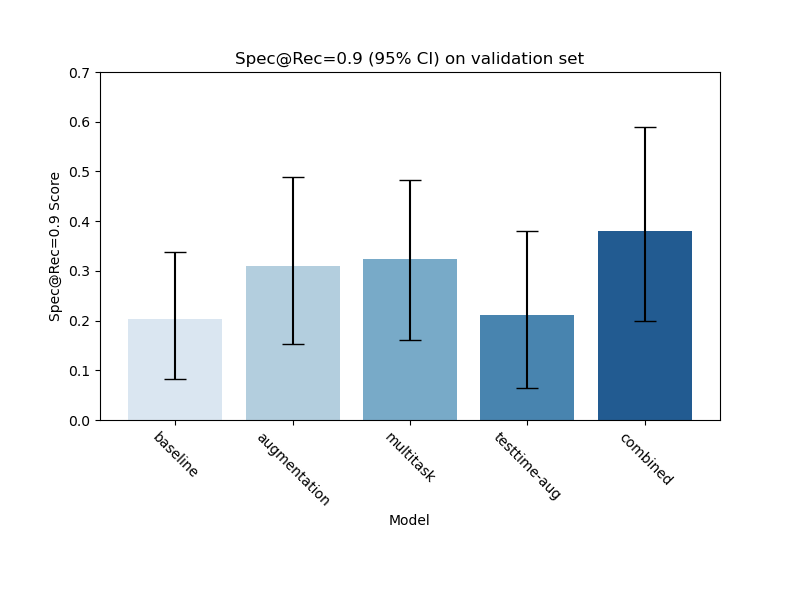}
    \end{subfigure}
    \begin{subfigure}{0.48\textwidth}
        \centering
        \includegraphics[trim=1.5cm 3.1cm 2.7cm 1.6cm, clip, width=\linewidth]{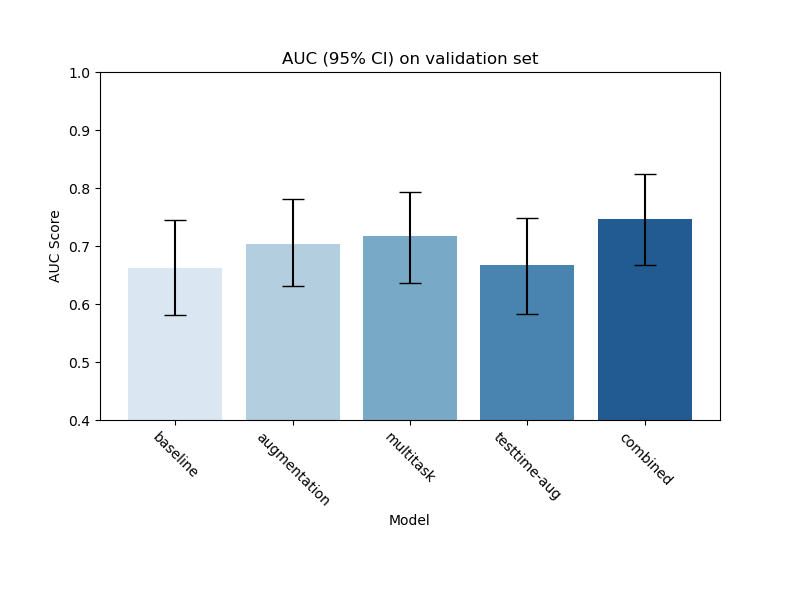}
    \end{subfigure}

    \vspace{1em}

    \begin{subfigure}{0.48\textwidth}
        \centering
        \includegraphics[trim=1.5cm 3.1cm 2.7cm 1.6cm, clip, width=\linewidth]{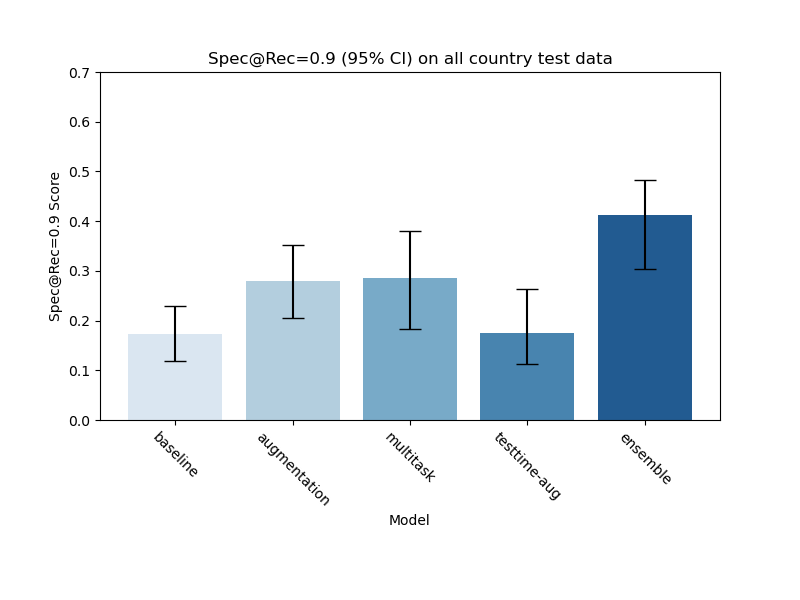}
        \label{fig:ablation-c}
    \end{subfigure}
    \begin{subfigure}{0.48\textwidth}
        \centering
        \includegraphics[trim=1.5cm 3.1cm 2.7cm 1.6cm, clip, width=\linewidth]{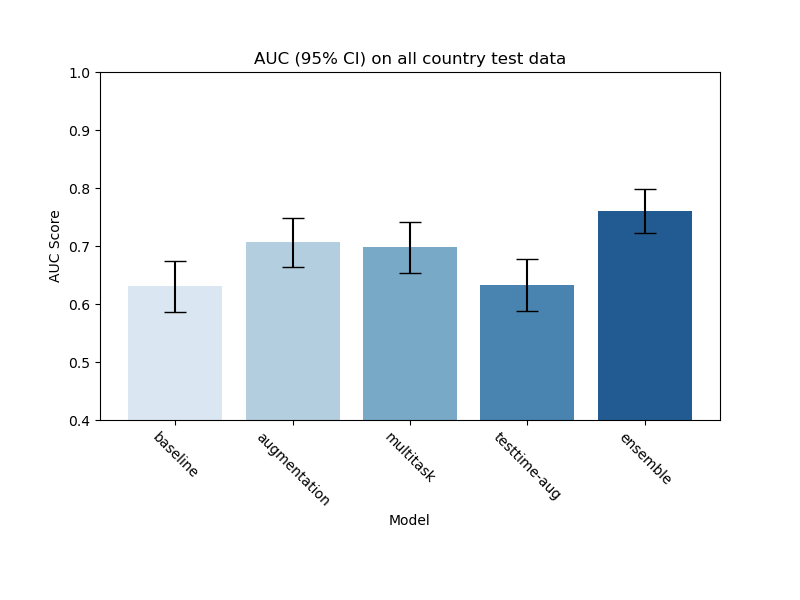}
        \label{fig:ablation-d}
    \end{subfigure}
    \caption{\textbf{Multi-task learning, heavy augmentations, and ensembling consistently improve performance on both validation and test sets.} (Left) Specificity@Sensitivity=0.9: Including the multi-task approach showcased the biggest performance boost on both validation set and on all country test data.
(Right) AUC: On our validation set, adding the multitask approach resulted in the highest increase in AUC, while adding augmentation gave the biggest boost in AUC score on all country test data.}
    \label{fig:ablation}
\end{figure}

 Ablation studies on \devval highlight multi-task learning as the most impactful design choice, increasing Spec@Sens=0.9 by 59.9\,pp and AUC by 8.3\,pp. Combining multi-task learning, heavy augmentation, and test-time augmentation yielded the highest overall scores, boosting Spec@Sens=0.9 by 88.1\,pp and AUC by 14.5\,pp.
When applied to the full test set, this trend persisted. Multi-task learning provided the largest single increase in Spec@Sens=0.9, improving it by 70.1\,pp. The final ensemble model achieved a maximum Spec@Sens=0.9 of 0.41, representing a 241\% increase over the baseline. Heavy augmentation had the greatest effect on AUC, increasing it by 12.6\,pp. The ensemble model reached the highest AUC of 0.76, a 20.6\,pp improvement over the baseline (see Fig. \ref{fig:ablation}).

\subsection{Q4: What characterizes images on which our method fails?}
\begin{figure}[!tbp]
    \centering
        \includegraphics[width=\linewidth]{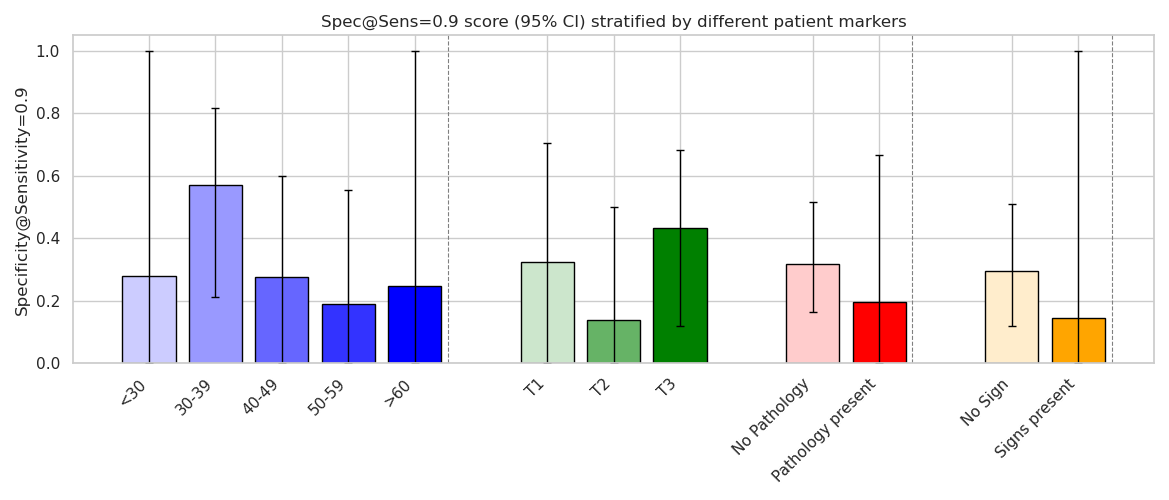}
        \includegraphics[width=\linewidth]{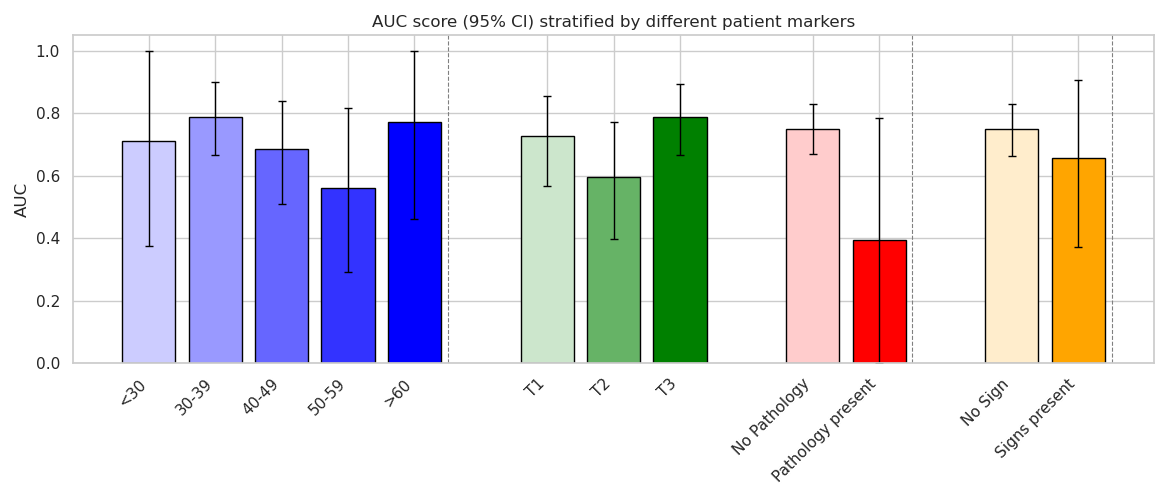}
    \caption{\textbf{The presence of other comorbidities had the biggest impact on performance.} Spec@Sens=0.9 (Top) and AUC score (Bottom) stratified by different patient markers are reported, as well as their respective bootstrapped 95\% CI which are represented as error bars. Among patient markers (Age, Transformation zone, Other Comorbidities and Pathognomonic signs), the presence of comorbidities have a clear impact on both metric scores, while the others show no consistent pattern.}
    \label{fig:failure}
\end{figure}

 Failure analysis was conducted by stratifying performance across four metadata categories described in section \ref{sec:data-curation}. Results are depicted in Fig. \ref{fig:failure}.

\paragraph{Age} Increasing age is often associated with a receding and less visible transformation zone\cite{desai2024squamocolumnar}, which complicates visual assessment during colposcopy. However, performance does not degrade consistently with age. AUC values ranged from 0.56 to 0.79, while Spec@Sens=0.9 scores ranged from 0.19 to 0.57.

\paragraph{Transformation Zone Type}
As the transformation zone becomes less visible from T1 to T3, classification becomes more challenging. However, performance remained variable without a clear trend. AUC scores ranged from 0.60 to 0.79, and Spec@Sens=0.9 values ranged from 0.14 to 0.43.

\paragraph{Presence of Other Pathologies}
In patients without comorbidities, the model achieved an AUC of 0.75. In contrast, cases with additional pathologies yielded an AUC of 0.40, indicating poor model performance. This represents an 87\,pp AUC performance reduction and a 38.2\,pp decrease in Spec@Sens=0.9. Due to the limited sample size of n=11, confidence intervals for this group are wide.

\paragraph{Presence of Pathognomonic Signs}
Pathognomonic signs are typically indicative of high-grade disease. Surprisingly, the presence of such signs reduced model performance. Cases without signs achieved an AUC of 0.75, while cases with signs dropped to 0.65, a 13.3\,pp decrease. Spec@Sens=0.9 fell by 51.4\,pp when signs were present.

\newpage
\section{Discussion}
\label{sec:discussion}
This study is the first to leverage data from multiple countries to develop and validate a DL algorithm for cervical lesion classification in colposcopy images.

 We present a novel DL method classifying cervical lesions into \lsil and \hsil using only acid-stain colposcopy images routinely available in clinical exams. Validated on datasets from four countries, this approach enables low-cost, scalable deployment in live clinical settings, supporting healthcare providers in the Global South with immediate, point-of-care decision support for biopsy or treatment.
Our method outperforms human experts in the challenging \lsil/\hsil classification task by achieving significantly higher sensitivity, the key metric for detecting all treatable lesions, even though humans demonstrate better specificity. With more data, performance could improve further, especially in difficult out-of-distribution scenarios where generalization remains a challenge.

A unique contribution of this work is the systematic investigation of failure cases to guide future improvements. Stratification by age showed no clear performance trend, contrary to expectations given age-related changes such as less visible transformation zones or increased comorbidities. Similarly, no performance difference was observed across transformation zone types, suggesting visibility does not impact classification. However, comorbidities were associated with worse model performance, likely due to obstructed regions of interest and the limited sample size of such cases. Notably, the presence of pathognomonic signs did not enhance classifier performance, possibly due to their complexity and underrepresentation during training, which may confuse the model.

 Our study would benefit from additional annotated data to increase the sample size for all lesion classes, improving the reliability of conclusions drawn from failure case analyses. Nevertheless, compared to prior work, this study is unique in its diversity and scale, incorporating datasets with varied ethnic backgrounds, imaging qualities, sensor types, and lighting conditions.

 Unlike previous studies which often rely on multi-modal inputs such as saline or iodine images or additional meta-data, our method uses only acid-stain colposcopy images to facilitate live, accessible screening in low-resource settings. Prior methods' code was not publicly available, requiring us to reimplement baselines.
To address domain shifts between medical centers, we explored self-supervised pretraining with SimCLR to learn domain-invariant features, focusing on semantic content rather than imaging artifacts. Initial results showed no performance gains. Similarly, leveraging general computer vision foundation models did not improve results, highlighting the challenges posed by the limited size of colposcopy datasets.

 Future research will expand training and test datasets across countries with varying image and class distributions. Increasing the representation of diverse comorbidities is crucial as these significantly impact model performance and clinical applicability. Improving model generalization remains a priority, with potential in domain generalization techniques such as feature alignment, disentanglement, meta-learning, self-supervised learning, ensembling, and distillation\cite{yoon2024domain}. 
Explainable AI methods including saliency maps\cite{dahan2025hybrid} and vision-language models\cite{gong2024revolutionizing} could enhance clinical trust and usability.

 Our study lays foundational work toward global AI-driven colposcopy, uniquely addressing lesion classification with multi-country data and aiming for translation to low- and middle-income countries. Future efforts should focus on generalization across diverse clinical settings, especially considering comorbidities, and promote cross-institutional data sharing to foster comparability and progress.

\section{Declarations}
\label{sec:declarations}

\subsection*{Declaration of Funding}
This work was supported by the Dieter Schwarz Stiftung gGmbH.

\subsection*{Declaration of generative AI and AI-assisted technologies}
During the preparation of this work the author used GPT-4 in order to improve grammar and help with rewording for conciseness. After using this tool, the author reviewed and edited the content as needed and takes full responsibility for the content of the published article.
\subsection*{Data Availability Statement}
The private data sets used in this study are not publicly available at this time as we are engaged in ongoing curation and development in collaboration with external partners. The data may be made available upon completion of these processes and subject to agreement with all involved parties. Public data is available upon request from the respective sources\cite{iarc-dataset, socolov2023annocerv}.
\clearpage
\appendix


\bibliographystyle{elsarticle-num-names} 
\bibliography{references}

@article{yoon2024domain,
  title={Domain generalization for medical image analysis: A review},
  author={Yoon, Jee Seok and Oh, Kwanseok and Shin, Yooseung and Mazurowski, Maciej A and Suk, Heung-Il},
  journal={Proceedings of the IEEE},
  year={2024},
  publisher={IEEE}
}

@article{dahan2025hybrid,
  title={A hybrid XAI-driven deep learning framework for robust GI tract disease diagnosis},
  author={Dahan, Fadl and Shah, Jamal Hussain and Saleem, Rabia and Hasnain, Muhammad and Afzal, Maira and Alfakih, Taha M},
  journal={Scientific Reports},
  volume={15},
  number={1},
  pages={21139},
  year={2025},
  publisher={Nature Publishing Group UK London}
}

@article{gong2024revolutionizing,
  title={Revolutionizing gastrointestinal endoscopy: the emerging role of large language models},
  author={Gong, Eun Jeong and Bang, Chang Seok},
  journal={Clinical endoscopy},
  volume={57},
  number={6},
  pages={759--762},
  year={2024},
  publisher={Korean Society of Gastrointestinal Endoscopy}
}

@misc{whomort,
  author       = {{World Health Organization}},
  title        = {Cervical cancer screening},
  year         = {2025},
  howpublished = {\url{https://www.who.int/data/gho/indicator-metadata-registry/
                       \\imr-details/3240}},
  note         = {Accessed: 2025-07-15}
}

@misc{iarc-dataset,
  author       = {{World Health Organization}},
  title        = {Cervical-Image-Bank},
  year         = {2025},
  howpublished = {\url{https://screening.iarc.fr/cervicalimagebank.php}},
  note         = {Accessed: 2025-07-15}
}

@misc{iarc-via,
  author       = {{World Health Organization}},
  title        = {IARC-VIA},
  year         = {2025},
  howpublished = {\url{https://prevention.datacollect.iarc.who.int/redcap/surveys/?s=P4DX7FFLC3}},
  note         = {Accessed: 2025-07-15}
}

@misc{iarc-colpobank,
  author       = {{World Health Organization}},
  title        = {IARC-Imagebank},
  year         = {2025},
  howpublished = {\url{https://prevention.datacollect.iarc.who.int/redcap/surveys/?s=7DKER78J7K}},
  note         = {Accessed: 2025-07-15}
}

@article{socolov2023annocerv,
  title={AnnoCerv: A new dataset for feature-driven and image-based automated colposcopy analysis},
  author={SOCOLOV, Razvan},
  journal={Acta Univ. Sapientiae Informatica},
  volume={15},
  number={2},
  pages={306--329},
  year={2023}
}

@inproceedings{eisenmann2023winner,
  title={Why is the winner the best?},
  author={Eisenmann, Matthias and Reinke, Annika and Weru, Vivienn and Tizabi, Minu D and Isensee, Fabian and Adler, Tim J and Ali, Sharib and Andrearczyk, Vincent and Aubreville, Marc and Baid, Ujjwal and others},
  booktitle={Proceedings of the IEEE/CVF conference on computer vision and Pattern recognition},
  pages={19955--19966},
  year={2023}
}

@article{wiesenfarth2021methods,
  title={Methods and open-source toolkit for analyzing and visualizing challenge results},
  author={Wiesenfarth, Manuel and Reinke, Annika and Landman, Bennett A and Eisenmann, Matthias and Saiz, Laura Aguilera and Cardoso, M Jorge and Maier-Hein, Lena and Kopp-Schneider, Annette},
  journal={Scientific reports},
  volume={11},
  number={1},
  pages={2369},
  year={2021},
  publisher={Nature Publishing Group UK London}
}

@article{maier2024metrics,
  title={Metrics reloaded: recommendations for image analysis validation},
  author={Maier-Hein, Lena and Reinke, Annika and Godau, Patrick and Tizabi, Minu D and Buettner, Florian and Christodoulou, Evangelia and Glocker, Ben and Isensee, Fabian and Kleesiek, Jens and Kozubek, Michal and others},
  journal={Nature methods},
  volume={21},
  number={2},
  pages={195--212},
  year={2024},
  publisher={Nature Publishing Group US New York}
}

@inproceedings{christodoulou2024confidence,
  title={Confidence intervals uncovered: Are we ready for real-world medical imaging AI?},
  author={Christodoulou, Evangelia and Reinke, Annika and Houhou, Rola and Kalinowski, Piotr and Erkan, Selen and Sudre, Carole H and Burgos, Ninon and Boutaj, Sofi{\`e}ne and Loizillon, Sophie and Solal, Ma{\"e}lys and others},
  booktitle={International Conference on Medical Image Computing and Computer-Assisted Intervention},
  pages={124--132},
  year={2024},
  organization={Springer}
}

@article{li2023segmentation,
  title={A segmentation model to detect cevical lesions based on machine learning of colposcopic images},
  author={Li, Zhen and Zeng, Chu-Mei and Dong, Yan-Gang and Cao, Ying and Yu, Li-Yao and Liu, Hui-Ying and Tian, Xun and Tian, Rui and Zhong, Chao-Yue and Zhao, Ting-Ting and others},
  journal={Heliyon},
  volume={9},
  number={11},
  year={2023},
  publisher={Elsevier}
}

@article{cho2020classification,
  title={Classification of cervical neoplasms on colposcopic photography using deep learning},
  author={Cho, Bum-Joo and Choi, Youn Jin and Lee, Myung-Je and Kim, Ju Han and Son, Ga-Hyun and Park, Sung-Ho and Kim, Hong-Bae and Joo, Yeon-Ji and Cho, Hye-Yon and Kyung, Min Sun and others},
  journal={Scientific reports},
  volume={10},
  number={1},
  pages={13652},
  year={2020},
  publisher={Nature Publishing Group UK London}
}

@article{chen2023application,
  title={Application of EfficientNet-B0 and GRU-based deep learning on classifying the colposcopy diagnosis of precancerous cervical lesions},
  author={Chen, Xiaoyue and Pu, Xiaowen and Chen, Zhirou and Li, Lanzhen and Zhao, Kong-Nan and Liu, Haichun and Zhu, Haiyan},
  journal={Cancer medicine},
  volume={12},
  number={7},
  pages={8690--8699},
  year={2023},
  publisher={Wiley Online Library}
}

@article{bai2022assessing,
  title={Assessing colposcopic accuracy for high-grade squamous intraepithelial lesion detection: a retrospective, cohort study},
  author={Bai, Anying and Wang, Jiaxu and Li, Qing and Seery, Samuel and Xue, Peng and Jiang, Yu},
  journal={BMC Women's Health},
  volume={22},
  number={1},
  pages={9},
  year={2022},
  publisher={Springer}
}

@article{wei2022improving,
  title={Improving colposcopic accuracy for cervical precancer detection: a retrospective multicenter study in China},
  author={Wei, Bingrui and Zhang, Bo and Xue, Peng and Seery, Samuel and Wang, Jiaxu and Li, Qing and Jiang, Yu and Qiao, Youlin},
  journal={BMC cancer},
  volume={22},
  number={1},
  pages={388},
  year={2022},
  publisher={Springer}
}

@article{yu2022segmentation,
  title={Segmentation of the cervical lesion region in colposcopic images based on deep learning},
  author={Yu, Hui and Fan, Yinuo and Ma, Huizhan and Zhang, Haifeng and Cao, Chengcheng and Yu, Xuyao and Sun, Jinglai and Cao, Yuzhen and Liu, Yuzhen},
  journal={Frontiers in Oncology},
  volume={12},
  pages={952847},
  year={2022}
}

@article{yan2021multi,
  title={Multi-state colposcopy image fusion for cervical precancerous lesion diagnosis using BF-CNN},
  author={Yan, Ling and Li, Shufeng and Guo, Yi and Ren, Peng and Song, Haoxuan and Yang, Jingjing and Shen, Xingfa},
  journal={Biomedical Signal Processing and Control},
  volume={68},
  pages={102700},
  year={2021},
  publisher={Elsevier}
}

@article{sha2024cervifusionnet,
  title={CerviFusionNet: A multi-modal, hybrid CNN-transformer-GRU model for enhanced cervical lesion multi-classification},
  author={Sha, Yuyang and Zhang, Qingyue and Zhai, Xiaobing and Hou, Menghui and Lu, Jingtao and Meng, Weiyu and Wang, Yuefei and Li, Kefeng and Ma, Jing},
  journal={iScience},
  volume={27},
  number={12},
  year={2024},
  publisher={Elsevier}
}

@article{yuan2020application,
  title={The application of deep learning based diagnostic system to cervical squamous intraepithelial lesions recognition in colposcopy images},
  author={Yuan, Chunnv and Yao, Yeli and Cheng, Bei and Cheng, Yifan and Li, Ying and Li, Yang and Liu, Xuechen and Cheng, Xiaodong and Xie, Xing and Wu, Jian and others},
  journal={Scientific reports},
  volume={10},
  number={1},
  pages={11639},
  year={2020},
  publisher={Nature Publishing Group UK London}
}

@article{li2024evaluation,
  title={Evaluation of the diagnostic performance of colposcopy in the detection of cervical high-grade squamous intraepithelial lesions among women with transformation zone type 3},
  author={Li, Xiaoxiao and Zhao, Yunzhi and Xiang, Fenfen and Zhang, Xinpei and Chen, Zixi and Zhang, Mengzhe and Kang, Xiangdong and Wu, Rong},
  journal={BMC cancer},
  volume={24},
  number={1},
  pages={381},
  year={2024},
  publisher={Springer}
}

@incollection{mello2023cervical,
  title={Cervical intraepithelial neoplasia},
  author={Mello, Vickie and Sundstrom, Renee K},
  booktitle={StatPearls [Internet]},
  year={2023},
  publisher={StatPearls Publishing}
}

@inproceedings{tan2019efficientnet,
  title={Efficientnet: Rethinking model scaling for convolutional neural networks},
  author={Tan, Mingxing and Le, Quoc},
  booktitle={International conference on machine learning},
  pages={6105--6114},
  year={2019},
  organization={PMLR}
}

@article{desai2024squamocolumnar,
  title={Squamocolumnar junction visibility, age, and implications for cervical cancer screening programs},
  author={Desai, Kanan T and Hansen, Natasha and Rodriguez, Ana-Cecilia and Befano, Brian and Egemen, Didem and Gage, Julia C and Wentzensen, Nicolas and Lopez, Catya and Jeronimo, Jose and de Sanjose, Silvia and others},
  journal={Preventive medicine},
  volume={180},
  pages={107881},
  year={2024},
  publisher={Elsevier}
}

@INPROCEEDINGS{5206848,
  author={Deng, Jia and Dong, Wei and Socher, Richard and Li, Li-Jia and Kai Li and Li Fei-Fei},
  booktitle={2009 IEEE Conference on Computer Vision and Pattern Recognition}, 
  title={ImageNet: A large-scale hierarchical image database}, 
  year={2009},
  volume={},
  number={},
  pages={248-255},
  doi={10.1109/CVPR.2009.5206848}}

\end{document}